%% file: main.tex
\pdfoutput=1
\documentclass[11pt]{article}
\usepackage{booktabs}
\usepackage{graphicx}
\usepackage{algpseudocode}
\usepackage{hyperref}
\usepackage[]{acl}
\usepackage{multirow}
\usepackage[inline]{enumitem}
\usepackage{enumerate}
\usepackage{amsfonts}
\usepackage{latexsym}
\usepackage{xspace}
\usepackage{amsmath}
\usepackage[T1]{fontenc}
\usepackage{pifont}
\usepackage{amssymb}
\usepackage{newfloat}
\usepackage{listings}
\usepackage{multirow}
\usepackage{booktabs}
\usepackage{times}
\usepackage[inline]{enumitem}
\usepackage{latexsym}
\usepackage[T1]{fontenc}
\usepackage[utf8]{inputenc}
\usepackage[inline]{enumitem}
\usepackage[dvipsnames]{colortbl}
\usepackage[utf8]{inputenc}
\usepackage[T1]{fontenc}
\usepackage{babel}
\usepackage[toc,page]{appendix}
\usepackage{makecell}
\usepackage{soul}
\usepackage{verbatim}
\usepackage{arydshln}
\usepackage{amssymb}
\usepackage{tcolorbox}
\usepackage{xcolor}
\usepackage{color}
\usepackage{colortbl}

\usepackage{graphicx} 
\usepackage[linesnumbered,ruled,vlined]{algorithm2e}
\usepackage{adjustbox}
\usepackage{footmisc}
\usepackage{microtype}
\usepackage{bbm}
\usepackage{wasysym}
\usepackage{tikz}
\usepackage{tcolorbox}
\usepackage{siunitx}
\usepackage{pifont}
\usepackage{longtable}
\usepackage{supertabular}
\usepackage[fixed]{fontawesome5}

\usepackage[utf8]{inputenc}

\usepackage{microtype}
\usepackage[nameinlink]{cleveref}
\usepackage{inconsolata}

\title{ Multi-agent Application System in Office Collaboration Scenarios}
\definecolor{Gainsboro}{rgb}{0.86, 0.86, 0.86}

\definecolor{Gray}{gray}{0.95}
\definecolor{LightCyan}{rgb}{0.88,1,1}

\newtcbox{\hlprimarytab}{on line, rounded corners, box align=base, colback=c3!10,colframe=white,size=fbox,arc=3pt, before upper=\strut, top=-2pt, bottom=-4pt, left=-2pt, right=-2pt, boxrule=0pt}
\newtcbox{\hlsecondarytab}{on line, box align=base, colback=red!10,colframe=white,size=fbox,arc=3pt, before upper=\strut, top=-2pt, bottom=-4pt, left=-2pt, right=-2pt, boxrule=0pt}

\author{
Songtao Sun$^{1}$\space\space\space
Jingyi Li$^1$\space\space\space
Yuanfei Dong$^1$\space\space\space
Haoguang Liu$^1$\space\space\space\\
{\bf
    Chenxin Xu$^1$\space\space\space
    Fuyang Li$^1$\space\space\space
    Qiang Liu$^1$
}\\
$^1$Kingsoft Office Software Inc., Wuhan, China \\
sunsongtao@wps.cn
}
\begin{document}
\maketitle
\begin{abstract}

This paper introduces a multi-agent application system designed to enhance office collaboration efficiency and work quality.
The system integrates artificial intelligence, machine learning, and natural language processing technologies, achieving functionalities such as task allocation, progress monitoring, and information sharing.
The agents within the system are capable of providing personalized collaboration support based on team members' needs and incorporate data analysis tools to improve decision-making quality.
The paper also proposes an intelligent agent architecture that separates Plan and Solver, and through techniques such as multi-turn query rewriting and business tool retrieval, it enhances the agent's multi-intent and multi-turn dialogue capabilities.
Furthermore, the paper details the design of tools and multi-turn dialogue in the context of office collaboration scenarios, and validates the system's effectiveness through experiments and evaluations.
Ultimately, the system has demonstrated outstanding performance in real business applications, particularly in query understanding, task planning, and tool calling.
Looking forward, the system is expected to play a more significant role in addressing complex interaction issues within dynamic environments and large-scale multi-agent systems.

\end{abstract}

\input{sections/01-introduction}

\input{sections/02-related-work}

\input{sections/03-methodology}
\input{sections/04-train-eval}

\input{sections/05-business}

\input{sections/06-conclusion}

\bibliography{custom}

\bigskip

\end{document}

%% file: sections/01-introduction.tex
\section{Introduction}\label{sec:intro}

Although large language models (LLMs) have achieved significant performance in a wide range of natural language processing tasks (Wang et al., 2023c; Chang et al., 2023), they still face some inherent limitations, such as outdated information (Qin et al., 2023b; Mallen et al., 2023).
Tool learning has been proposed to endow LLMs with various auxiliary resources, such as search engines (Shi et al., 2024b; Nakano et al., 2021) or calculators (Schick et al., 2023; Gao et al., 2023), which enable them to act as intelligent agents using tools and enhance their ability to solve specific complex tasks.
Most previous studies allowed LLM-based agents to sequentially perform multiple actions with tools in a predefined order (Yao et al., 2023; Yang et al., 2023b; Zhuang et al., 2023).
The agent first decomposes the task and gradually plans a series of tools.
For each step, the agent executes the tool by passing parameters and continuously integrates useful intermediate results into the next action prediction.

\begin{figure}[t]
        \centering
	\includegraphics[width=1\linewidth]{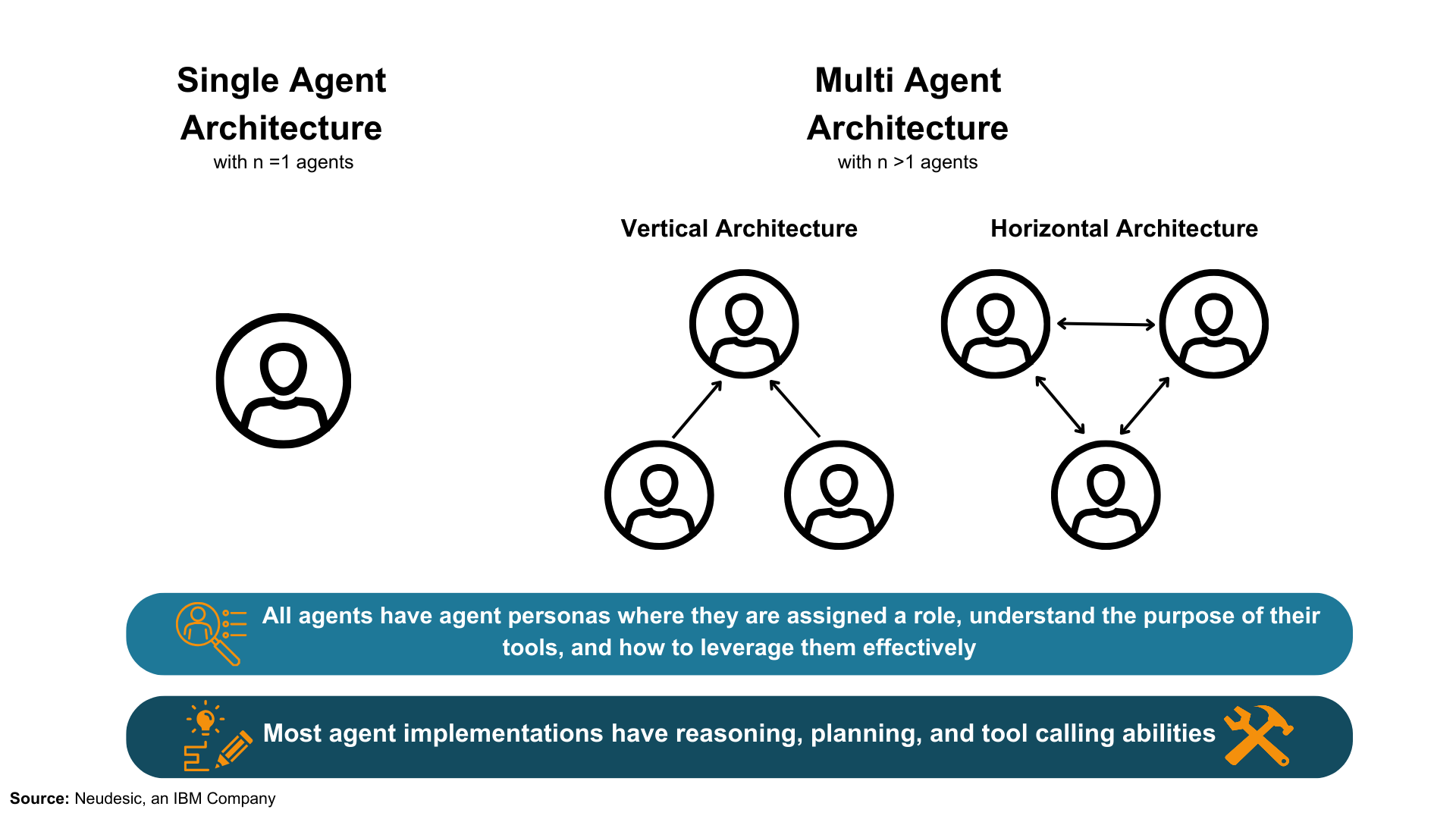}
        \caption{A Visualization of Single and Multi-agent Architectures~\citep{masterman2024landscape}.}
 \label{fig:intro}
\end{figure}

Although existing methods have made progress, they face two challenges in practice.
Firstly, most of them alternate between planning and execution through predefined processes (Yang et al., 2023b; Song et al., 2023), which inevitably limits their flexibility when dealing with frequently occurring anomalous errors in tool usage workflows (Shi et al., 2024a; Wang et al., 2023b; Prasad et al., 2023). When tools cannot be invoked, it is crucial for the agent to be able to correct its erroneous behavior, rather than directly moving to the next step immediately after the erroneous response.
Secondly, adapting a single LLM-based agent to learn to perform various specialized actions for solving tasks is challenging (Dziri et al., 2023; Yin et al., 2023).
Solving real-world tasks involves a diverse set of actions with significant differences, such as planning, execution, and reflection, which draw on different aspects of LLMs (Shen et al., 2024; Qiao et al., 2024).
Therefore, developing effective agent processes and organizing tool usage models to solve real-world tasks remains a challenging research topic.

This paper constructs a multi-agent application system in the context of office collaboration scenarios, aiming to enhance team collaboration efficiency and work quality by integrating various intelligent technologies.
The system utilizes artificial intelligence, machine learning, and natural language processing technologies to achieve task allocation, progress monitoring, and information sharing functions. The intelligent agents within the system can provide personalized collaboration support based on the needs and work characteristics of team members.
In addition, the system integrates advanced data analysis tools to help teams better understand workflow and improve decision quality.
In this way, team members can communicate and collaborate more efficiently, thus maintaining a leading position in the competitive business environment.

In our in-depth research and development of office collaboration scenarios, we have achieved the following significant contributions:

(1) We successfully built a multi-agent application system, specifically designed for the complex needs of office collaboration.
By adopting a multi-agent collaborative architecture with a master-slave pattern (Liu et al., 2023; Sun et al., 2023), we integrated the intelligent agent architecture of Plan and Solver (Schick et al., 2023; Qiao et al., 2024), which enables the system to efficiently process various business tasks.
In addition, we integrated business tool retrieval recall technology and Query multi-turn rewriting technology, the combination of which allows the system to support multi-intent recognition and multi-turn dialogue functions, greatly enhancing the efficiency and smoothness of office collaboration.
We also paid special attention to user experience, optimizing the user interface and interaction design, making the system more intuitive and user-friendly, allowing users to complete tasks more conveniently.

(2) Aiming at the complexity of tool invocation in office collaboration scenarios and the need for multi-turn tool connections, we designed a series of innovative tool invocation strategies and connection mechanisms.
To support the implementation of these strategies and mechanisms, we developed a multi-turn data generation framework capable of simulating actual office collaboration dialogue processes and generating high-quality training data.
At the same time, to ensure the accuracy and practicality of the data, we also formulated a comprehensive multi-turn data annotation scheme, ensuring the quality of the data and the reliability of the system through manual annotation.
We also introduced machine learning algorithms to conduct in-depth data analysis, further improving the accuracy and response speed of tool invocation.

(3) After experimental evaluation and validation in real business scenarios, our system demonstrated outstanding performance and stability.
Through comparative experiments, we proved that our system is more efficient and has a lower error rate than traditional systems when dealing with complex business processes.
In addition, we collected user feedback, continuously optimizing system functions to meet the ever-changing needs of office collaboration.

%% file: sections/02-related-work.tex
\begin{figure*}[t!]
    \centering
\includegraphics[width=1\linewidth]{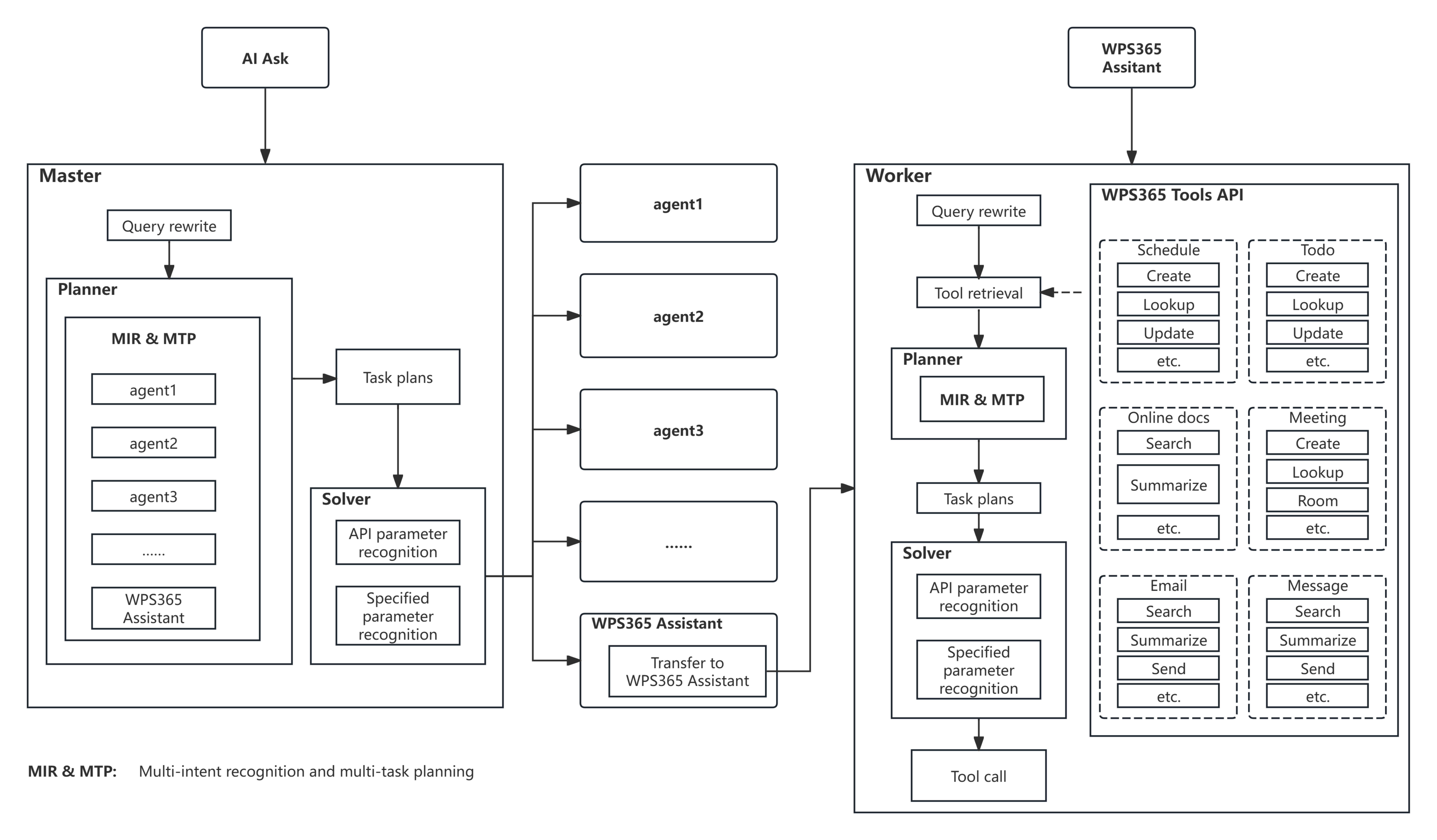}
    \caption{System Architecture.}
\label{fig:architecture}
\end{figure*}

\section{Related Work}\label{sec:related-work}

\paragraph{LLMs tool learning.}
The use of external tools to enhance LLMs has been proven to be a promising approach for solving practical tasks (Bran et al., 2023; Qu et al., 2024; Wang et al., 2024b).
Previous work has typically empowered tool-learning agents through supervised fine-tuning (Patil et al., 2023; Yang et al., 2023a; Gao et al., 2024) or prompt learning (Lu et al., 2023; Shen et al., 2023). Specifically, the former trains LLMs on tool usage datasets, guiding LLMs on how to use tools from data (Wang et al., 2023c; Hao et al., 2023).
The latter directly demonstrates tool usage methods to LLMs using context examples (Paranjape et al., 2023; Kim et al., 2023).
However, using tools to solve complex tasks involves various actions, such as deciding which tools to use, passing which parameters, and how to utilize the results (Schick et al., 2023; Qiao et al., 2024).
Forcing a single agent to learn all these capabilities can put more pressure on it (Yin et al., 2023; Prasad et al., 2023).
Furthermore, as tasks become more complex, LLM-based agents struggle to integrate lengthy task-solving contexts due to their limited working memory when predicting the next action (Shi et al., 2023).
In contrast, we propose utilizing an Agent architecture with separated Plan and Solver components and enhance the agent's multi-intent and multi-turn dialogue capabilities by introducing techniques such as multi-turn query rewriting and business tool recall.

\paragraph{Multi-agent Collaboration.}
The collaboration of multiple agents has demonstrated strong performance across various tasks (Liu et al., 2023; Sun et al., 2023; Zhang et al., 2023), enhancing the capabilities of individual agents (Talebirad and Nadiri, 2023; Mohtashami et al., 2023; Qian et al., 2023).
Recent research has introduced multiple agents into debates, engaging in discussions for a fixed number of rounds (Wang et al., 2023a; Liang et al., 2023), which has improved their factuality (Cohen et al., 2023) and reasoning abilities (Du et al., 2023; Fu et al., 2023).
In tool learning tasks, recent work has employed different agents to achieve task planning and execution separately, thereby reducing the workload of individual agents (Shen et al., 2024; Song et al., 2023; Qiao et al., 2024).
Despite their progress, their agent processes have been simplified to predefined procedures (Prasad et al., 2023), and they face difficulties in handling frequent exceptional error handling in tool usage workflows (Zhuang et al., 2023; Wang et al., 2023b).
In our work, we propose the use of a master-slave pattern multi-agent collaborative framework to coordinate multiple specialized single agents, generating solutions through agent collaboration, and improving the task completion rate in business scenarios.

%% file: sections/03-methodology.tex
\section{Methodology}\label{sec:method}

\begin{figure*}[htb]
    \centering
\includegraphics[width=1\linewidth]{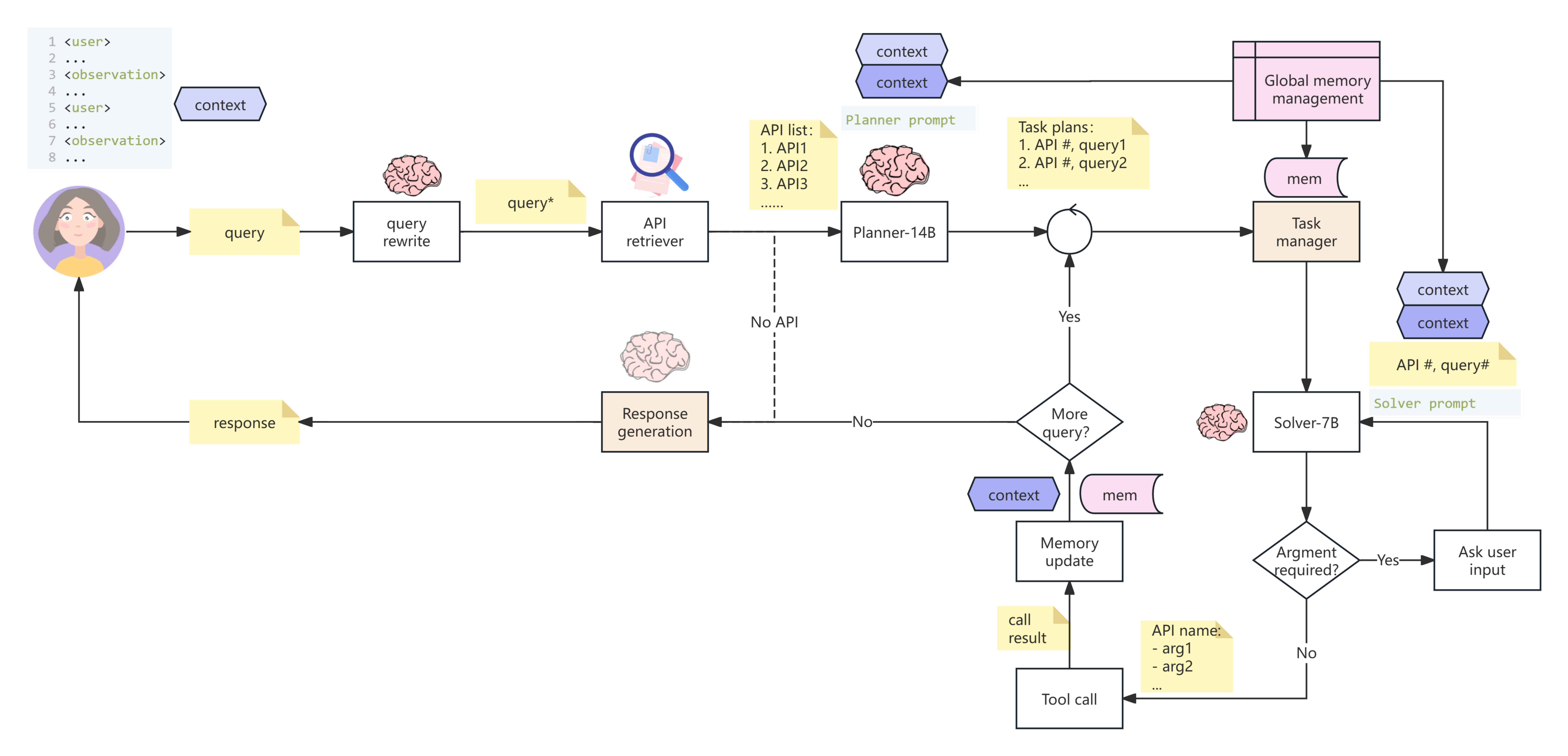}
    \caption{Planner and Solver Flowchart.}
\label{fig:planner-solver}
\end{figure*}
\subsection{Overall System}\label{sec:system}

The WPS AI Enterprise Edition consists of three parts: AI Hub, AI Docs, and Copilot Pro.
Its core value lies in helping enterprises build an "Enterprise Brain," providing "out-of-the-box" AI capabilities, allowing employees to utilize AI within a controlled corporate environment, thereby enhancing production efficiency.
Among these, Copilot Pro (also known as the Smart Assistant) includes product modules such as AI Ask, WPS365 Assistant, and Customizable Smart Assistant.
The WPS365 Assistant primarily offers out-of-the-box AI capabilities for the WPS365 suite to enterprises.

As in Figure~\ref{fig:architecture}, this paper utilizes a master-slave pattern multi-agent collaborative architecture as the top-level system architecture, with the underlying master-slave nodes adopting a Plan+Solver agent architecture.
It also integrates Query multi-turn rewriting technology and tool retrieval techniques to construct an intelligent assistant system with various office capabilities.

\input{table/03-office-tool-design}

\subsection{Key Technologies}
\paragraph{Master-slave mode multi-agent collaborative architecture.}\label{sec:master-slave}

In this architecture, AI Ask acts as the master node, undertaking core functions such as user interaction, task distribution, memory management, and agent coordination.
Concurrently, some other agents (like chitchat bots, text-to-image generation, online searching, etc.) and WPS365 assistant as worker nodes, each executing specific tasks.
These tasks can be performed by simple agents (e.g., chitchat bots and text-to-image generation) or by complex agents (such as online searching and WPS365 assistant).

\paragraph{Planner+Solver agent architecture.}\label{sec:Planner+Solver}

As Figure~\ref{fig:planner-solver} shows, the Plan+Solver agent architecture consists of two models: the Planner Model and the Solver Model.

Planner Model: The primary responsibility is to decompose the rewritten user query into multiple sub-tasks and utilize the context information from the multi-turn dialogue to reasonably plan a sequence of sub-tasks that satisfy the user's complex requirements.

Solver Model: Primarily responsible for parameter identification of sub-tasks, by analyzing the context content of multi-turn dialogues, accurately extract the parameter values required for the specified API or the parameter values of specified parameters.

\paragraph{Tool retrieval.}\label{sec:retrieval}
In the Master node, the number of tools involved is relatively small, allowing all of them to be directly placed into the Prompt to complete the Plan work.
However, in the Worker node, the volume of tool data can reach dozens, and the length of tool description texts is usually large (due to the large number of parameters), thus it is necessary to select the most relevant multiple tools rather than all tools to be placed into the Prompt based on the query.

\paragraph{Query rewrite.}\label{sec:rewrite}
\begin{figure}[t]
    \centering
\includegraphics[width=1\linewidth]{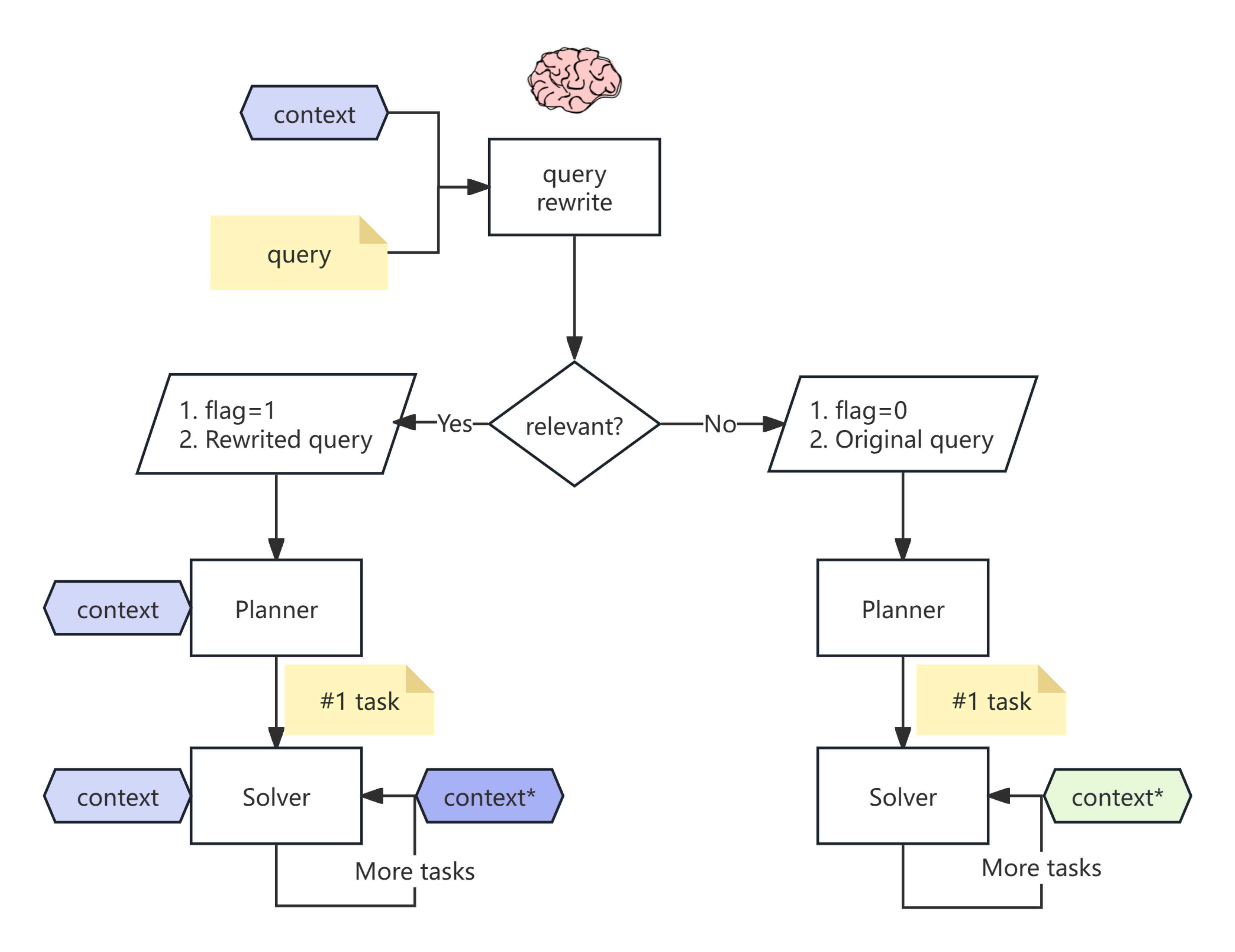}
    \caption{Query Rewrite Flowchart.}
\label{fig:rewrite}
\end{figure}

In response to the challenges posed by colloquial query expressions in business scenarios (including phenomena such as deixis and ellipsis), this technology employs strategies such as context-related discrimination, query rewriting(shows in Figure~\ref{fig:rewrite}), and intent distribution, and works in conjunction with a memory management mechanism, aiming to enhance the agent's ability to handle multi-turn conversations.

\input{table/03-office-tool-transition}

\begin{figure*}[ht]
\centering
\includegraphics[width=1\linewidth]{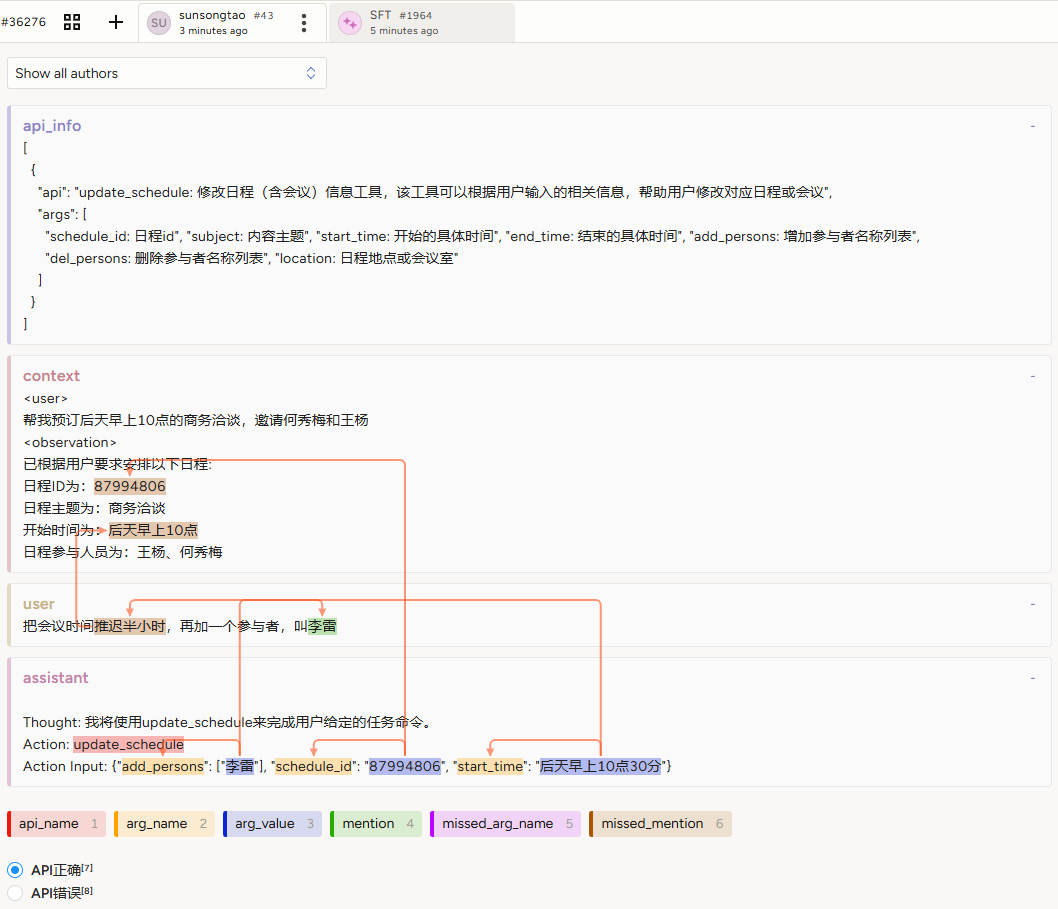}
\caption{Data Annotation Interface ().}
\label{fig:data-annotation}
\end{figure*}

\subsection{Office collaboration scenarios}\label{sec:office}

This section mainly introduces the API design of related information tools in the office collaboration scenario, as well as the solution for multi-tool invocation during multi-round dialogues.

\paragraph{Tool designs.}\label{sec:tool}

WPS 365 offers collaborative features through the form of 365 native APIs for downstream invocation, which can achieve the scheduling of native API functions through the integration of AI capabilities. For the six common scenarios in office collaboration, the following business tool APIs have been designed and packaged as in Table~\ref{tab:tool-design}.

\paragraph{Multi-turn dialogue design.}\label{sec:dialogue}

To support multiple tool invocations during multi-turn dialogue processes, it is necessary to enhance the inter-turn contextual awareness.
Through analysis and research, this paper ultimately adopts a method of customizing tool invocations for the previous and current turns, and has designed the following schemes for email, schedule, to-do, and chat scenarios (shows in Table~\ref{tab:tool-transition}).
The formula for calculating the number of parameter combinations is $(2^m-1) \times (2^n-1)$, where $m$ is the number of parameters for the previous turn tool, and $n$ is the number of parameters for the current turn tool.

\subsection{Data Engineering work}\label{sec:engineering}

This section mainly introduces the engineering work of data generation and annotation.

\paragraph{Multi-turn Data Generation Framework.}

Based on the characteristics of multi-round dialogue data in office collaboration tasks, a framework for a multi-agent workflow is defined for the generation of dialogue data samples.
Introducing multi-agent workflows, by defining roles through different modules (Flow, Task, Role, Agent), to utilize the Agent concept for construction.
Clarify the inheritance relationship between multiple workflow units: the chain of Flow -> Task -> Role -> Agent.

\paragraph{Multi-turn Data Annotation Scheme.}

Based on the characteristics of data from multiple rounds of tool calls, a labeling scheme suitable for multiple rounds has been developed on the Label Studio platform.

Figure~\ref{fig:data-annotation} presents a multi-turn dialogue sample that has been fully annotated, containing three main annotation categories: label, choice, and relation. A detailed description follows.
Within the label category, there are multiple tags used to annotate the correct information of APIs and their parameters: the api\_name tag is used to identify the correct API name; the arg\_name tag is used to identify the correct parameter name; the arg\_value tag is used to identify the correct parameter value; and the mention tag is used to identify the paragraph in the text where the correct parameter value is mentioned.
Additionally, there are several tags used to annotate missing information: the missed\_arg\_name tag is used to identify missing parameter names; the missed\_arg\_value tag is used to identify missing parameter values, primarily for backfilling enumerated parameter values; and the missed\_mention tag is used to identify missing query mention text segments.
In the choice category, there are two tags used to identify the correctness of API selection: the API correct tag is used to indicate that the selected API is correct; and the API error tag is used to indicate that the selected API is incorrect.
Finally, in the relation category, there is a related tag used to identify the correlation between different tags, i.e., their associative relationships.

%% file: table/03-office-tool-design.tex
\begin{table*}[th!]
\begin{tabular}{p{0.1\textwidth} p{0.2\textwidth} p{0.1\textwidth} p{0.5\textwidth}}
\toprule
\textbf{Scenario} & \textbf{API} & \textbf{Number of args} & \textbf{Function Description} \\
\midrule
Email & search\_email & 11 & Search emails based on input criteria. \\
 & send\_email & 5 & Send emails to specified recipients. \\
 & summary\_email & 2 & Generate summaries for emails or email content. \\
\hline
Schedule & create\_schedule & 5 & Create schedules or meetings with input details. \\
 & update\_schedule & 7 & Update schedules or meetings with input details. \\
 & find\_schedule\_status & 3 & Check availability of participants or personal schedules. \\
 & delete\_schedule & 1 & Cancel schedules or meetings. \\
\hline
Meeting & create\_meeting & 5 & Create meetings using create\_schedule. \\
 & find\_meetings & 2 & Search meetings based on time range. \\
 & find\_meeting\_room & 5 & Search meeting rooms with input details. \\
\hline
Chat & search\_chatmsg & 7 & Search chat messages with input criteria. \\
 & send\_chatmsg & 3 & Send chat messages and @mention users. \\
 & withdraw\_chatmsg & 2 & Withdraw specific chat messages. \\
 & summary\_chatmsg & 1 & Summarize chat messages by IDs. \\
 & search\_group\_chat & 1 & Search group chats by keywords. \\
 & find\_recent\_chat\_list & 2 & View recent chat sessions by time. \\
\hline
Todo & create\_todo & 2 & Create todo items with input details. \\
 & find\_todo & 3 & Query and filter todo items. \\
 & delete\_todo & 1 & Delete specified todo items. \\
\hline
Online & search\_files & 13 & Search cloud files with filtering options. \\
Documents & summary\_files & 1 & Summarize cloud documents by IDs. \\
\bottomrule
\end{tabular}
\caption{Tool Designs for Office Collaboration Scenarios}
\label{tab:tool-design}
\end{table*}

%% file: table/03-office-tool-transition.tex
\begin{table*}[th!]
\begin{center}
\begin{tabular}{c c c c c}
\toprule
\textbf{Scenario} & \textbf{Previous Tool} & \textbf{Current Tool} & \textbf{\# Combinations} & \textbf{Complexity} \\
\midrule
Email & \text{search\_email} & \text{summary\_email} & \text{6141} & \text{High} \\
& \text{summary\_email} & \text{send\_email} & \text{93} & \text{Medium} \\
\hline
Schedule & \text{create\_schedule} & \text{update\_schedule} & \text{3937} & \text{High} \\
& \text{create\_schedule} & \text{delete\_schedule} & \text{31} & \text{Low} \\
& \text{update\_schedule} & \text{delete\_schedule} & \text{127} & \text{Medium} \\
& \text{find\_schedule\_status} & \text{create\_schedule} & \text{217} & \text{Medium} \\
& \text{find\_schedule\_status} & \text{delete\_schedule} & \text{7} & \text{Low} \\
& \text{find\_schedule\_status} & \text{update\_schedule} & \text{889} & \text{High} \\
\hline
Todo & \text{find\_todo} & \text{delete\_todo} & \text{7} & \text{Low} \\
& \text{create\_todo} & \text{delete\_todo} & \text{3} & \text{Low} \\
\hline
Chat & \text{search\_chatmsg} & \text{summary\_chatmsg} & \text{127} & \text{Medium} \\
& \text{search\_group\_chat} & \text{summary\_chatmsg} & \text{1} & \text{Low} \\
& \text{find\_recent\_chat\_list} & \text{summary\_chatmsg} & \text{3} & \text{Low} \\
& \text{summary\_chatmsg} & \text{send\_chatmsg} & \text{7} & \text{Low} \\
& \text{search\_chatmsg} & \text{withdraw\_chatmsg} & \text{381} & \text{High} \\
& \text{search\_group\_chat} & \text{send\_chatmsg} & \text{7} & \text{Low} \\
\bottomrule
\end{tabular}
\caption{Tool Transition Scenarios with Corresponding Parameter Combinations and Complexity}
\label{tab:tool-transition}
\end{center}
\end{table*}

%% file: sections/04-train-eval.tex
\section{Training and Evaluation}\label{sec:experiment}

\subsection{Models and datasets}
\textbf{Models.} In the system constructed in this article, query rewriting, Planner, and Solver all utilize the Qwen2.5-Instruct series as the base model for full-parameter SFT.
After experimental comparisons in the early stages, query rewriting ultimately selected Qwen2.5-3B-Instruct, Planner opted for Qwen2.5-14B-Instruct, and Solver chose Qwen2.5-7B-Instruct, which have been deployed and run online.
For tool recall, the bge-m3 model of FlagEmbedding was selected for full-parameter fine-tuning.

\textbf{Datasets.} The training set samples for the query rewriting model cover daily general scenarios, AI Ask scenarios, and the full WPS365 scenarios, with the final dataset reaching a scale of 10K.
The training set samples for the Planner model also cover the aforementioned three scenarios and include 10 auxiliary tools, requiring samples to encompass both single-intent and multi-intent situations, with the final dataset reaching a scale of 20K.
The training samples for the Solver model also span the three scenarios mentioned above, but without the addition of auxiliary tools.
Additionally, samples with designated parameter extraction are included, with the final dataset reaching a scale of 30K.
The tool recall vector model is tailored for the WPS365 full-scenario API, with a small portion of daily common data added for balance.
Like the Planner model, it includes both single-intent and multi-intent samples, thereby constructing positive and negative example pairs, with the final dataset reaching a scale of 15K.

\subsection{Evaluation results}

\paragraph{Query Rewrite Evaluation.}
The query rewriting model simultaneously accomplishes context relevance judgment and rewriting tasks.
The former uses the $relate_acc$ metric, while the latter employs common ROUGE and BLEU metrics in addition to using LLM for ground-truth rewriting semantic consistency evaluation (running 3 times).
The table below lists the evaluation results of two versions of fine-tuned models, where the ROUGE and BLEU metrics focus on the comparison of textual literals, thus using the ground-truth metric as the standard for measuring rewriting quality.

\paragraph{Tool recall evaluation.}
The evaluation metrics for the tool recall model use top-n recall rate, employing two different negative sampling strategies, and are evaluated on single-intent and multi-intent test datasets.
Specifically, v1 constructs the training set by treating clauses in the multi-intent training set as single sentences, training only on business data; v2 constructs training data with multiple positive examples for multi-intent data samples, and the negative examples are randomly selected from the remaining APIs, with 3 chosen at random.

\input{table/04-evaluation-results-1}

\paragraph{Planner evaluation.}
The Planner model is designed using the ReWoo data paradigm, which plans a query into multiple sub-tasks, essentially multiple clause texts.
The ROUGE metric is used to measure the completeness of information in each clause, $sub_tasks_num-acc$ measures the accuracy of sentence splitting quantity, and API-acc measures the accuracy of the API for single-step intent.
The following four versions of datasets are constructed for evaluation:
• v1: A mixed dataset of 28,648 single-intent and multi-intent data;
• v2: A mixed dataset of 34,116 single-intent and multi-intent data, with 5,468 planning auxiliary data;
• v3: A mixed dataset of 15,960 single-intent and multi-intent data, with unchanged planning auxiliary data;
• v4: A mixed dataset of 35,960 single-intent and multi-intent data, with unchanged planning auxiliary data, and an additional 10,000 ShareGPT data.

\paragraph{Solver evaluation.}
The Solver model output includes both API and parameters.
Previous studies typically calculated indicators separately, whereas this paper adopts a strict accuracy measure, considering the output correct only when both parts are consistent.
During the development process, three versions of training datasets were constructed and evaluated on both the business test set and the context test set.
Notably, the business side conducted comparative evaluations between version v1 and two outstanding closed-source LLMs, GLM4 and GPT4, demonstrating the advantages of the fine-tuned model.

\input{table/04-evaluation-results-2}

%% file: table/04-evaluation-results-1.tex
\begin{table*}[t!]
\centering
\begin{tabular}{c c c c c c}
\toprule
\textbf{Base Model} & \textbf{Data Version} & \textbf{relate\_acc}
& \textbf{ROUGE} & \textbf{BLEU} & \textbf{ground-truth} \\
\midrule
Qwen2.5-3B-Instruct & v1 & 0.9523 & 0.7802 & 0.8413 & 0.79±0.03 \\
Qwen2.5-3B-Instruct & v2 & 0.9877 & 0.7769 & 0.8431 & 0.81±0.03 \\
\bottomrule
\end{tabular}
\caption{Evaluation Results of Query Rewrite Models.}
\label{tab:eval-rewrite}
\end{table*}

\begin{table*}[t!]
\centering
\begin{tabular}{c c  c c  c c}
\toprule
\multirow{2}{*}{\textbf{Base Model}} & \multirow{2}{*}{\textbf{Data Version}} &
\multicolumn{2}{|c}{\textbf{Single-Intent Test Set}} & \multicolumn{2}{|c}{\textbf{Multi-intent Test Set}} \\
 &  & \textbf{top3} & \textbf{top5} & \textbf{top3} & \textbf{top5} \\
\midrule
bge-m3 & v1 & 0.8934 & 0.9492 & 0.8629 & 0.9036 \\
bge-m3 & v2 & 0.9898 & 1.0 & 0.8883 & 0.9594 \\
\bottomrule
\end{tabular}
\caption{Evaluation Results of Tool Recall Models.}
\label{tab:eval-recall}
\end{table*}

%% file: table/04-evaluation-results-2.tex
\begin{table*}[t!]
\centering
\begin{tabular}{c c c c c}
\toprule
\textbf{Base Model} & \textbf{Data Version} & \textbf{ROUGE} & \textbf{sub\_tasks\_num-acc} & \textbf{API-acc} \\
\midrule
Qwen2.5-14B-Instruct & v1 & 0.8178 & 0.8579 & 0.9892 \\
Qwen2.5-14B-Instruct & v2 & 0.8107 & 0.8477 & 0.9874 \\
Qwen2.5-14B-Instruct & v3 & 0.8216 & 0.8528 & 0.9796 \\
Qwen2.5-14B-Instruct & v4 & 0.8159 & 0.8528 & 0.9892 \\
\bottomrule
\end{tabular}
\caption{Evaluation Results of Planner Models.}
\label{tab:eval-planner}
\end{table*}

\begin{table*}[t!]
\centering
\begin{tabular}{p{0.25\textwidth} p{0.1\textwidth} p{0.1\textwidth} p{0.1\textwidth} p{0.1\textwidth} p{0.1\textwidth}}
\toprule
\textbf{Base Model} & \textbf{Data Version} & \textbf{Total Data} &
\textbf{Business Evaluation} & \textbf{Business Test Set} & \textbf{Context Test Set} \\
\midrule
GLM4 & - & - & 0.42 & - & - \\
GPT4 & - & - & 0.48 & - & - \\
Qwen2.5-7B-Instruct & v1 & 19547 & 0.678 & 0.8366 & 0.95215 \\
Qwen2.5-7B-Instruct & v2 & 34054 & - & 0.9085 & 0.85646 \\
Qwen2.5-7B-Instruct & v3 & 26478 & - & 0.89634 & 0.9195 \\
\bottomrule
\end{tabular}
\caption{Evaluation Results of Solver Models.}
\label{tab:eval-solver}
\end{table*}

%% file: sections/05-business.tex
\section{Actual Business Results}\label{sec:result}

In terms of actual business effectiveness, WPS365 Assistant and AI Ask modules have been successfully applied in various corporate environments.
These modules not only support office collaboration scenarios such as email, schedule, meetings, chat, to-do items, and cloud documents, but also support the processing of colloquial queries through multi-turn dialogue design.

\begin{figure}
    \centering
\includegraphics[width=1\linewidth]{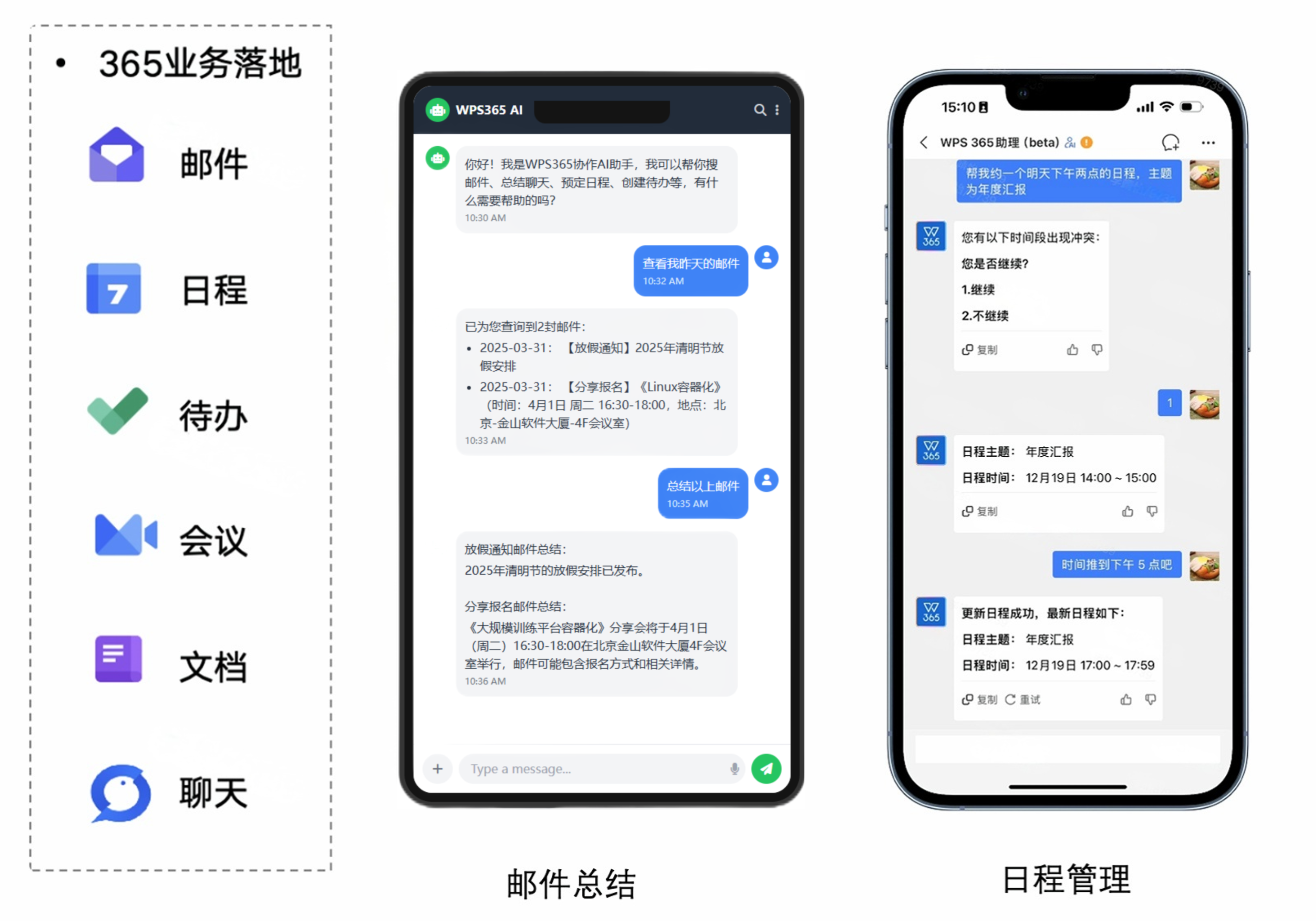}
    \caption{Business Results of the System.}
\label{fig:biz-results}
\end{figure}

For instance, in the email scenario, it supports operations such as searching, summarizing, and sending emails, and possesses the capability for multi-turn conversations, capable of handling colloquial requests; in the schedule scenario, it supports the entire process of checking availability, creating, updating, and deleting schedules, supports multi-turn conversations, and can handle colloquial requests.

%% file: sections/06-conclusion.tex
\section{Conclusions}\label{sec:conclusion}

In this era of information explosion, the efficiency and quality of corporate office collaboration are directly related to the competitiveness and market position of the enterprise. To meet this challenge, this article introduces an innovative multi-agent application system.
By integrating artificial intelligence, machine learning, and natural language processing technologies, the system achieves key functions such as task allocation, progress monitoring, and information sharing.
The agents within the system can provide personalized collaboration support based on the needs and work characteristics of team members, thereby greatly enhancing the efficiency and quality of team collaboration.

The core of the system lies in its unique intelligent agent architecture that separates the Planner and Solver.
This architecture allows the system to decouple decision-making during task planning and execution from actual operations when dealing with complex business tasks.
In this way, the system can respond more flexibly to various business scenarios, especially excelling in handling tasks that require multi-step, multi-intention interactions.

To further enhance the multi-intent and multi-turn dialogue capabilities of agents, this paper proposes and implements query multi-turn rewriting and business tool retrieval techniques.
The combination of these techniques enables the system to better understand users' consecutive requests and maintain the coherence of context in multi-turn dialogues.
For example, when handling email scenarios, the system is not only capable of understanding users' requests to search for emails but can also summarize or forward emails based on further instructions from the user.
This capability is particularly important in practical business applications, as it significantly reduces the repetitive work and communication costs for users when completing tasks.

In future development, the system is expected to play a greater role in dealing with complex interaction issues in dynamic changing environments and large-scale multi-agent systems.
With continuous technological progress and optimization, the system will better serve corporate office collaboration, helping enterprises enhance their competitiveness.
Through continuous innovation and improvement, we believe that the system will become an important tool in the field of corporate office collaboration, bringing a more efficient and intelligent way of working to enterprises.

\input{table/05-business-results}

%% file: table/05-business-results.tex
\begin{table*}[t!]
\begin{center}
\begin{tabular}{ p{0.08\textwidth} p{0.18\textwidth} p{0.32\textwidth} p{0.32\textwidth}}
\toprule
\textbf{Scenario} & \textbf{Task} & \textbf{Query} & \textbf{Execution Effect} \\
\midrule
Email & Search Email & Search for the emails I received today & Precisely recognize the intent to search for emails and complete the query for the specified email \\

& Summarize Email & Summarize the emails I received today & Able to follow the context, then summarize the content of the found emails. \\

& Search Email & Forward the emails received today to Jiashu Xia & Support email forwarding (permission not open yet) \\

& Send Email & Send an email to Jiashu Xia, content: Salaries for December have been issued, please check! & Support email sending (permission not open yet) \\

\hline

Schedule & Create Schedule & Create a meeting at 3 PM today, the topic is project discussion, invite Jiashu Xia & Support the creation of schedules (meetings) \\

& Update Schedule & Move the start time up to 2 PM & Support updating the schedule created or updated in the last turn \\

& Update Schedule & Update the meeting at 3 PM today, change the topic to product discussion & Support updating information for the specified schedule (meeting) \\

& Delete Schedule & Delete all meetings at 3 PM today & Support deleting the specified schedule (meeting) \\

& Check Free Time & Check Jiashu Xia's free time tomorrow? & Support querying the free time of the specified person \\
\bottomrule
\end{tabular}
\caption{Business Results of the System.}
\label{tab:biz-results}
\end{center}
\end{table*}